\documentclass[conference]{IEEEtran}
\usepackage{blindtext, graphicx}
\usepackage{hyperref}
\usepackage{float}
\usepackage{amsmath}
\hypersetup{
    colorlinks=true,
    linkcolor=blue,
    filecolor=magenta,      
    urlcolor=cyan,
}
\ifCLASSINFOpdf
\else
\fi

\begin{document}
%
\title{Automatic salt deposits segmentation: A deep learning approach}

\author{
\IEEEauthorblockN{Mikhail Karchevskiy}
\IEEEauthorblockA{Aurteen Inc.\\
Novosibirsk, Russia\\
Email: karchevskiy.michael@gmail.com}
\and
\IEEEauthorblockN{Insaf Ashrapov}
\IEEEauthorblockA{Tele2 Russia\\
Moscow, Russia\\
Email: iashrapov@gmail.com}
\and
\IEEEauthorblockN{Leonid Kozinkin}
\IEEEauthorblockA{Upwork Global Inc.\\
Novosibirsk, Russia\\
Email: l.kozinkin@gmail.com}
}

\maketitle

\begin{abstract}

One of the most important applications of seismic reflection is the hydrocarbon exploration which is closely related to salt deposits analysis. This problem is very important even nowadays due to it's non-linear nature. Taking into account the recent developments in deep learning networks TGS-NOPEC Geophysical Company hosted the Kaggle competition for salt deposits segmentation problem in seismic image data. In this paper, we demonstrate the great performance of several novel deep learning techniques merged into a single neural network which achieved the 27th place (top 1\%) in the mentioned competition. Using a U-Net with ResNeXt-50 encoder pre-trained on ImageNet as our base architecture, we implemented Spatial-Channel Squeeze \& Excitation, Lovasz loss, CoordConv and Hypercolumn methods. The source code for our solution is made publicly available at \url{https://github.com/K-Mike/Automatic-salt-deposits-segmentation}.

\end{abstract}

\begin{IEEEkeywords}
Deep Learning, Image Segmentation, Computer Vision, Reflection Seismology
\end{IEEEkeywords}

\IEEEpeerreviewmaketitle

\section{Introduction}

The salt deposits seismic analysis problem was known more than hundred years ago and even influenced development of the reflection seismic method \cite{telford_1976}. The salt analysis is considered especially interesting due to the close contact with hydrocarbon deposits which leads to additional problems in the exploring and extraction process \cite{andresen2011hydrocarbon}. 
Since the texture of salt deposits is rather chaotic the salt segmentation problem is complicated and still very important nowadays \cite{jones2014seismic}. The first approach to this problem was the manual seismic images interpretation by geophysics specialists. Over the years there were developed some mathematical methods to automate this process \cite{halpert2008salt, hegazy2014texture}, however, the accuracy of those, especially in some complex cases, was not sufficient, thus some hybrid methods were presented \cite{wu2016methods}.

The recent developments in deep learning methods provided a huge impact in such chaotic data analysis areas as geoscience and greatly increased identification accuracy \cite{waldeland2018convolutional}. This led to the current Kaggle competition hosted by TGS (the world’s leading geoscience data company) \cite{kaggle_salt} which aim is to build an algorithm that automatically and accurately identifies if a subsurface target is salt or not.

\section{Model Architecture}

\begin{figure}[ht]
\centering
\includegraphics[width=6cm]{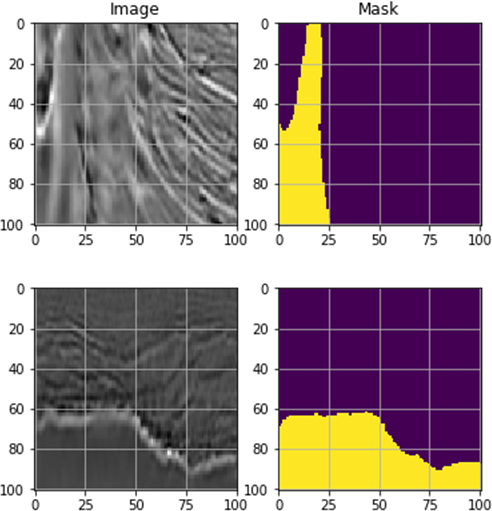}
\caption{Samples the of given dataset: seismic images and true masks.}
\label{fig::data_sample}
\end{figure}

Seismic reflection data was provided as $101 \times 101$ pixel images and binary masks of salt deposits for the train dataset (see Fig. \ref{fig::data_sample}). Total train and test datasets consisted of 8000 and 18000 images respectively. Additionally, a depth was provided for each image.

The predicitons were evaluated on the mean average precision at different intersection over union (IoU) thresholds:

$$T = \{0.5, 0.55, 0.6, \dots, 0.9, 0.95\}$$

At each threshold value $t$, a precision value is calculated based on the number of true positives ($\mathit{TP}$), false negatives ($\mathit{FN}$), and false positives ($\mathit{FP}$) resulting from comparing the predicted object to all ground truth objects. The final score is calculated as

$$M = \frac 1{|T|} \sum_{t \in T} {\frac{\mathit{TP}(t)}{\mathit{TP}(t) + \mathit{FP}(t) + \mathit{FN}(t)}}$$

To increase the train dataset we used some augmentation techniques \cite{buslaev2018albumentations}. The most positive impact was provided by the horizontal flip method. Also, some improvements were achieved by using brightness manipulations, horizontal shifts and rotations. However, due to the nature of the seismic data some valuable information was encoded in the height component of images, thus, vertical flips or big rotations decreased the overall accuracy. As the final solution we developed a single 5-fold model with horizontal flip TTA (test time augmentation).

\begin{figure*}
\includegraphics[width=\textwidth]{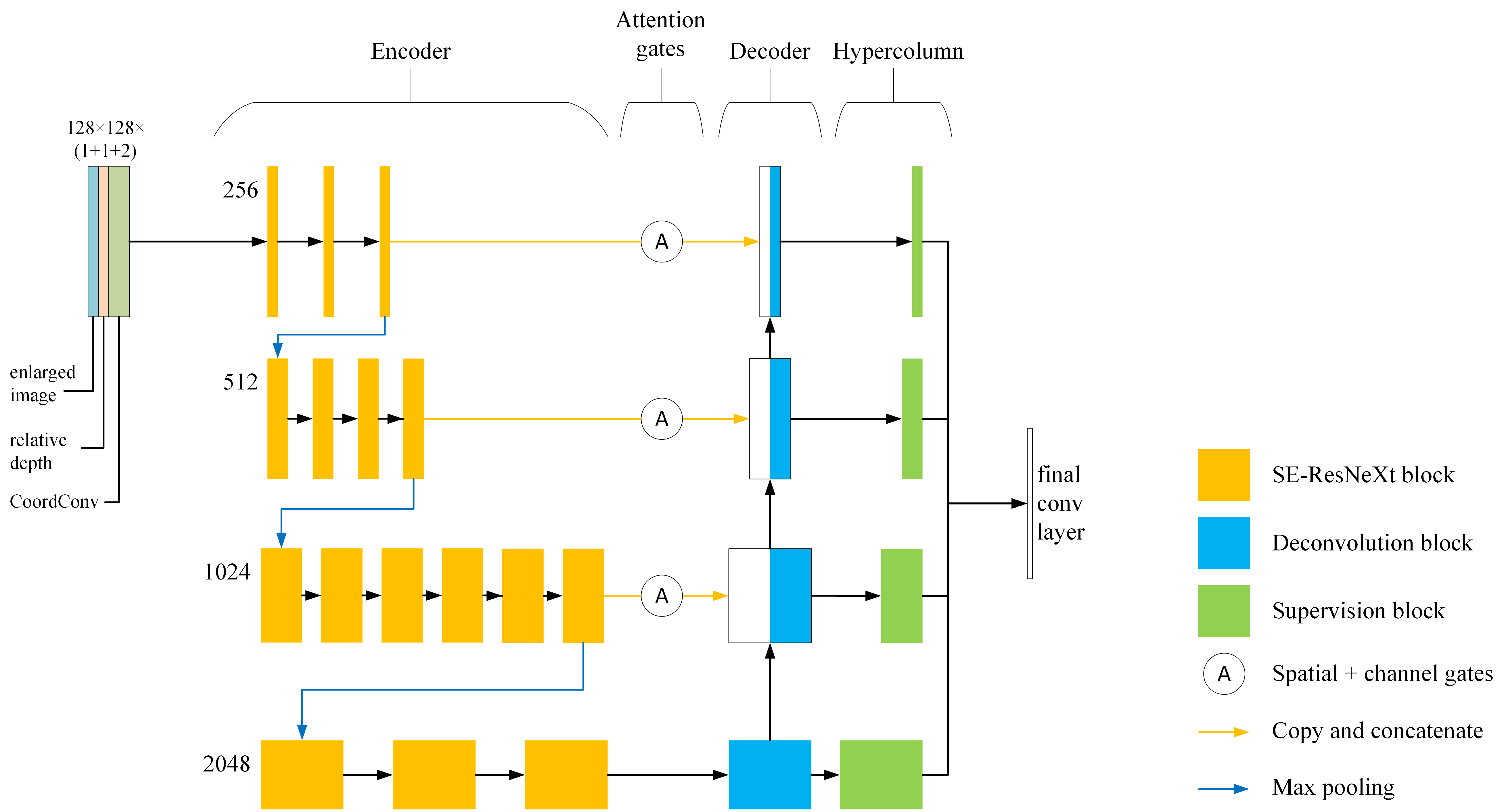}
\caption{Proposed model structure: the input images are enlarged from $101 \times 101$ pixels to $128 \times 128$ pixels, the number of channels get increased from $1$ to $4$ by adding a relative depth channel and $2$ CoordConv channels; then the images get processed through the pre-trained encoder - ResNeXt bottlenecks (grouped in $3$, $4$, $6$, $3$ blocks connected via max pooling layers) with an Squeeze-and-Excitation block inside each bottleneck (orange); the decoder consists of $4$ consecutive deconvolution blocks (blue), each block processing information from a previous block, a corresponding encoder block and attention gates (spatial and channel-wise); each decoder is also connected to a deep supervision block (basically, convolution + upsample layers) implementing the Hypercolumn technique (green); finally, all of the DSV outputs are concatenated to the final convolution layer.}
\label{fig::architecture}
\end{figure*}

Input grayscale images were extended to $128 \times 128$ pixel size by the reflect transformation and concatenated with additional channel-like data: relative depth

$$D_{rel}=\frac{D_i - D_{min}}{D_{max} - D_{min}}$$
as the 2nd channel and CoordConv \cite{liu2018intriguing} as the 3rd channel.

As our base model the popular approach using U-Net with pretrained custom encoder \cite{iglovikov2018ternausnet, shvets2018automatic, shvets2018angiodysplasia} was implemented. The model was further improved with SE-ResNeXt-50 encoder and modified first layer (CNN stride set to 1). It showed better results than other ResNet-like architectures and even outperformed SE-ResNeXt-101. Both encoder and decoder of our neural network were implemented using the ScSE (Spatial-Channel Squeeze and Excitation) method \cite{hu2017squeeze}. For further improvements the spatial and channel attention gates \cite{chen2017sca} were included in each encoder and decoder block. To get additional information from the CNN layers the hypercolumn representation technique \cite{hariharan2015hypercolumns} was used. Dropouts were removed from the model to significantly reduce the training time. The number of CNN filters was set to 32 (16 provided much less accuracy and 64 required sufficiently more computational resources). The complete model structure is described in Fig. \ref{fig::architecture}.

\begin{figure}[ht]
\centering
\includegraphics[width=9cm]{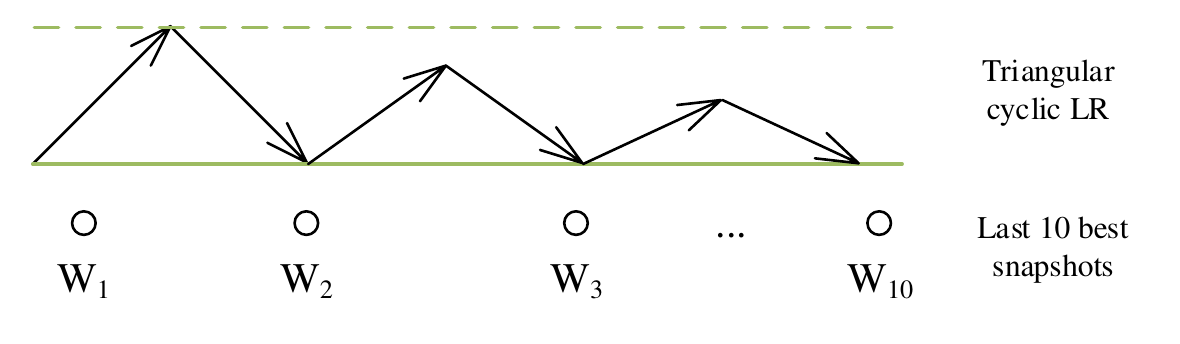}
\caption{Cyclic learning rate and snapshot ensembling.}
\label{fig::learning}
\end{figure}

For our model training the Adam optimizer was used, with batch size = 20, cyclic learning rate (triangular2 policy) and heavy snapshot ensembling, i.e. exponentially weighted average of last 10 best models (see in Fig. \ref{fig::learning}):

Let the snapshots $W_1, W_2, \dots, W_{10}$ be the last best snapshots, i.e. satisfying the condition

$$IoU(W_1) < IoU(W_2) < \dots < IoU(W_{10})$$

Then the ensemble should be calculated by formula:

\begin{figure*}
\includegraphics[width=\textwidth]{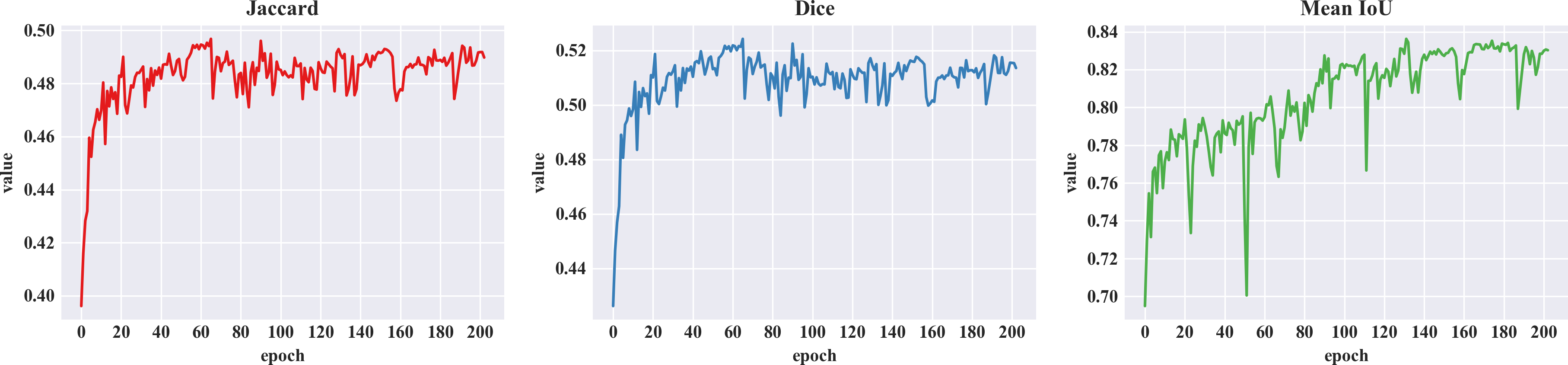}
\caption{The final model metric scores on the validation set for one fold.}
\label{fig::metrics}
\end{figure*}

\begin{multline} \nonumber
W_{ens} = \alpha \left[ W_{10} + (1 - \alpha) W_9 + (1 - \alpha)^2 W_8 + \dots \right. \\ \left. \dots + (1 - \alpha)^8 W_2  \right] + (1 - \alpha)^9 W_1, \;\;\; \alpha = \frac12
\end{multline}

This technique made models blending practically useless (the accuracy gain from using less successful models wasn't noticeable).

The model was trained for 80 epochs with BCE loss first and then continued with 0.1 BCE + 0.9 Lovasz loss \cite{matthew2018lovasz} until 50 epochs early stopping criteria reached. The Lovasz loss function was selected due to it's better performance for the IoU metrics optimization task (i.e. an improved segmentation on the edges of objects). The solution was coded using the PyTorch framework \cite{pytorch}. To refine the predicted masks small connected areas (both black and white) were removed by using the OpenCV \cite{opencv} function “cv2.connectedComponentsWithStats”.
Using higher binarization threshold ($thr>0.4$) seems to give a better score on the private validation dataset.

\section{Results}

To process the model described above we had a single NVidia GTX 1080 Ti GPU at our disposal. The full training and prediction cycle took about 24 hours. The metrics scores results achieved on the validation during the training process are shown on Fig. \ref{fig::metrics}. The proposed model was created and evaluated in several stages. The following table presents the dynamic of the public leaderboard score increase percentage for each implemented or discarded feature (starting from a general U-Net architecture):

\begin{center}
\begin{tabular}{ |l|c| } 
 \hline
Feature & Scores increase \\
\hline
\hline
 Vanilla U-Net & -- \\ 
 \hline
+  ResNet152 encoder & 1.26\% \\ 
 \hline
+  batch normalization & 0.25\% \\ 
 \hline
-- center layer max pooling & 0.62\% \\ 
--  dropout layers & \\ 
 \hline
+  Lovasz hinge loss & 0.62\% \\ 
 \hline
--  max pooling for all layers & 1.60\% \\ 
+  KFold validation & \\ 
 \hline
+  Hypercolumn & 0.85\% \\ 
 \hline
-  ResNet152 encoder & 0.36\% \\ 
+  SE-ResNeXt50 encoder & \\ 
 \hline
+  spatial/channel squeeze \& excitation & 0.72\% \\ 
 \hline
+  max pooling with stride = 1 & 0.24\% \\ 
 \hline
+  spatial/channel attention gates & 2.25\% \\ 
+  cyclic learning rate & \\ 
+  snapshots ensembling & \\ 
 \hline
+  TTA horizontal flip & 0.12\% \\ 
+  postprocessing & \\ 
 \hline
\end{tabular}
\end{center}

There were also other features which have been implemented but turned out useless and were discarded in the final model:

\begin{itemize}

\item Higher resolution input images: provided no noticeable improvements and caused much slower learning.

\item Dropout layers: slightly reduced the final scores and required more training time.

\item Jigsaw mosaics post-processing \cite{paikin2015solving}: produced no impact on the private validation dataset.

\item Morphology post-processing: affected both outliers and true masks, thus no positive gain achieved.

\end{itemize}

\section{Conclusion}

In our approach to the stated problem we showed the high efficiency of the deep learning methods. The predictions provided even by a single DL model were able to achieve the 27th place. Several novel techniques like CoordConv or Squeeze-and-Excitation networks showed great performance in real-world problems as well as ResNeXt-like architectures. Additionally, there were some optimizations and tuning tricks presented. Our solution is available as an open source project under MIT licence at \url{https://github.com/K-Mike/Automatic-salt-deposits-segmentation}.

\section*{Acknowledgment}
The authors would like to thank Open Data Science community \cite{ods} for many valuable discussions and educational help
in the growing field of machine/deep learning.

\ifCLASSOPTIONcaptionsoff
  \newpage
\fi

\bibliography{references}

\end{document}